\documentclass[conference]{IEEEtran}
\IEEEoverridecommandlockouts
\usepackage{cite}
\usepackage{amsmath,amssymb,amsfonts}
\usepackage{algorithmic}
\usepackage{graphicx}
\usepackage{textcomp}
\usepackage{xcolor}
\def\BibTeX{{\rm B\kern-.05em{\sc i\kern-.025em b}\kern-.08em
    T\kern-.1667em\lower.7ex\hbox{E}\kern-.125emX}}
\begin{document}

\title{Advancements in Myocardial Infarction Detection and Classification Using Wearable Devices: A Comprehensive Review\\
{\footnotesize \textsuperscript{}
}
\thanks{.}
}
\author{\IEEEauthorblockN{ Abhijith S\IEEEauthorrefmark{1}}
\IEEEauthorblockA{\textit{Dept. of Electronics and Comm.Engg.} \\
\textit{College of Engineering}\\
Trivandrum,India \\
tve20ae005@cet.ac.in}
\and
\IEEEauthorblockN{Arjun Rajesh\IEEEauthorrefmark{1}}
\IEEEauthorblockA{\textit{Dept. of Electronics and Comm.Engg.} \\
\textit{College of Engineering}\\
Trivandrum,India\\
tve20ae019@cet.ac.in}
\and
\IEEEauthorblockN{ Mansi Manoj\IEEEauthorrefmark{1}}
\IEEEauthorblockA{\textit{Dept. of Electronics and Comm.Engg.} \\
\textit{College of Engineering}\\
Trivandrum,India \\
tve20ae038@cet.ac.in}
\and
\IEEEauthorblockN{ Sandra Davis Kollannur\IEEEauthorrefmark{1}}
\IEEEauthorblockA{\textit{Dept. of Electronics and Comm.Engg.} \\
\textit{College of Engineering}\\
Trivandrum,India\\
tve20ae063@cet.ac.in}
\and
\IEEEauthorblockN{ Sujitta R V\IEEEauthorrefmark{1}}
\IEEEauthorblockA{\textit{Dept. of Electronics and Comm.Engg.} \\
\textit{College of Engineering}\\
Trivandrum,India\\
tve22ecra16@cet.ac.in}
\and
\IEEEauthorblockN{Jerrin Thomas Panachakel}
\IEEEauthorblockA{\textit{Dept. of Electronics and Comm.Engg.} \\
\textit{College of Engineering}\\
Trivandrum,India \\
jerrin.panachakel@cet.ac.in}
\and
\IEEEauthorblockA{\IEEEauthorrefmark{1} These authors contributed equally}
}

\maketitle

\begin{abstract}
Myocardial infarction (MI), commonly known as a heart attack, is a critical health condition caused by restricted blood flow to the heart. Early-stage detection through continuous ECG monitoring is essential to minimize irreversible damage. This review explores advancements in MI classification methodologies for wearable devices, emphasizing their potential in real-time monitoring and early diagnosis. It critically examines traditional approaches, such as morphological filtering and wavelet decomposition, alongside cutting-edge techniques, including Convolutional Neural Networks (CNNs) and VLSI-based methods. By synthesizing findings on machine learning, deep learning, and hardware innovations, this paper highlights their strengths, limitations, and future prospects. The integration of these techniques into wearable devices offers promising avenues for efficient, accurate, and energy-aware MI detection, paving the way for next-generation wearable healthcare solutions.
\end{abstract}

\begin{IEEEkeywords}
Myocardial infarction(MI),classification, wearables
\end{IEEEkeywords}

\section{Introduction}
Myocardial infarction (MI), also known as a heart attack, is caused by reduced blood flow to the heart chambers. MI can be silent and  undetected, or it can have serious effects and lead to death. Most myocardial infarctions are caused by coronary artery disease. When a coronary artery blockage occurs, there is a lack of oxygen within the heart muscle. Prolonged lack of oxygen supply to the heart can lead to  death and necrosis of myocardial cells. Patients experience chest discomfort or tightness that can spread to the neck, jaw, shoulders, or arms. MI is one of the major causes of passings causing 9 million fatalities annually around the world, which is anticipated to extend up to 12 million by 2030[10]. MI advances in three stages, specifically, early MI (EMI), intense MI (AMI) and constant MI (CMI) characterized by its seriousness. Delay in taking ECG may cause irreversible damage, so it's important to take ECGs in certain intervals to get an accurate result so necessary action can be taken as soon as possible. Many solutions have been proposed by researchers around the world for the analyzation and classification of MI. Wearable devices have recently taken the spotlight for being easily accessible, accurate, and user-friendly. Such devices also have the merit of being energy-efficient and comparatively less complex in their ministrations, which are cardinal factors to consider while designing ECG classification gadgets. In order to examine the operational capabilities of the devised contraption, it will have to be tested against existing data. 

 The following sections will delve deep into the comparisons between various MI classification methods put forward by researchers over the years, which will facilitate a clear understanding on the same.The paper explores various methodologies including machine learning,deep learning, VLSI, and IoT-based methods contributing to efficient and accurate detection and classification of Myocardial infarction that can be implemented in wearables for a timely analysis.By synthesizing findings from relevant studies, the review highlights strengths, limitations, and potential future directions in leveraging ECG signals for accurate and efficient myocardial stage identification.
\begin{table*}[h]
    \centering
     \caption{Comparison of different classification methods}
    \begin{tabular}{|c|c|l|l|}
        \hline
        \textbf{Serial  No}&\textbf{Reference} & \textbf{Classification Method}& \textbf{Accuracy}\\
        \hline
        1 & D. Sopic et al.[1]&Two level SVM Classifier& 90\%\\
        \hline
        2 & M. Odema et al [2]&BCNN& 91.22\%\\
        \hline
        3 & N. Rashid et al [3]&BCNN& 90.29\%\\
        \hline
        4 & D. Sopic et al [5]&Two level Hierachiacal classifiation& 83.26\%\\
        \hline
        5 & Yu-Hung Chuang et al [6]&MLP Classififer& 99.72\%\\
        \hline
        6 &    X. Ma et al [7]&CDD net& 98. 95\%\\
        \hline
        7 & Hadjem et al [8]&Statistical method& 73\%\\\hline
      
    \end{tabular}
    \label{tab:mytable}

\end{table*}
\section{Methodology}
\subsection{Preprocessing and Feature Extraction}
Feature extraction plays a crucial role in the process of myocardial infarction
(MI) detection,it involves identifying relevant patterns or characteristics from 
the physiological signals obtained through wearables. The goal is to transform 
raw data into a set of distinctive markers that can be analyzed to identify 
potential indicators of myocardial infarction. 

[1] and [3] converge on a shared traditional signal 
processing approach, incorporating morphological filtering, FIR band-pass 
filtering, and PanTompkin’s algorithm. These methods collectively address 
baseline wander, high-frequency noise, and R-peak detection for subsequent 
feature extraction. In parallel, [2] distinguishes itself by introducing 
advanced Convolutional Neural Networks (CNNs) and Binary CNNs, leveraging their
inherent convolution process for more efficient feature extraction. 

Expanding the repertoire, [5] employs wavelet decomposition but extends 
the feature set beyond conventional parameters. Normalized signal energy, 
fractal dimension, and various entropies are incorporated, broadening the scope
of information extracted from the ECG signals. Innovative strategies are evident
in  [7] and [8]. [7] explores the transformation of ECG signals 
into images using Hilbert curve mapping, introducing a unique encoding approach. 
On the other hand, [8] combines wavelet decomposition, baseline wandering 
removal, and the CUSUM algorithm for ST elevation analysis, presenting a 
holistic preprocessing strategy for MI detection.

Furthermore,[9] and [10] contribute to the discussion on feature 
selection. While [9] explores Relief, Minimal-Redundancy-Maximal Relevance, and 
Least Absolute Shrinkage and Selection Operator algorithms, [10] introduces an 
optimized ECG feature extraction algorithm based on derivatives. These 
approaches highlight the importance of selecting relevant features for accurate MI classification.
\begin{figure*}[h!]
    \centering
    \includegraphics[width=0.7\linewidth]{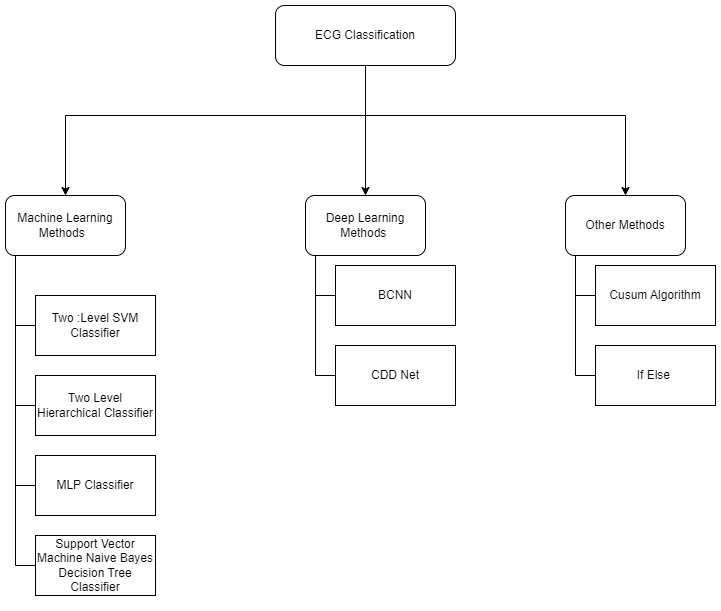}
    \caption{Tree diagram of different classification methods}
    \label{fig:enter-label}
\end{figure*}

\subsection{Classification Methods}

Classification models use various algorithms to analyze extracted features from
physiological signals and classify individuals into categories, distinguishing 
between those with and without indications of myocardial infarction. 
  
In the realm of wearable devices for myocardial infarction (MI) classification, various classification methods have been employed by numerous researchers and may fall under machine learning, deep learning, or other categories as illustrated in Fig.1. [1],  adopted a two-stage classification scheme where the first level, computationally efficient but less accurate, activates a more complex second-level classifier if the first-level classifier is unable to classify the sample data with the required confidence. Another prevalent method is the use of Convolutional Neural Networks (CNNs), exemplified in [2] and [3], where Binary Convolutional Neural Networks (BCNNs) are designed for memory efficiency, with additional mentions of SVM and Random Forest in related works.The proposed methodology in [2] incorporates Neural Architecture Search (NAS) using Multi-Objective Bayesian Optimization (MOBO) to optimize the design of neural network architectures for MI detection. [5] used a hiearchial classification with first stage being a binary classifier to classify between abnormal and normal ECG beats.Then in the second stage a multi-class classifier is used to classify the abnormal beats into one of four categories: MI, bundle branch block, premature ventricular contraction, or other. The paper also mentions that the classifiers were trained using a support vector machine (SVM) with a radial basis function (RBF) kernel.

In contrast, [9] explored multiple machine learning algorithms, including Support Vector Machines, Naive Bayes, and Decision Tree Classifier, for MI detection.  [7] employed a unique Convolutional Dendritic model (CDD) designed for feature extraction and logical relationship interpretation. [6] opted for a multilayer perceptron (MLP) classifier for MI classification based on derived vectorcardiography, demonstrating its capability to categorize ECG beats into specific MI classes. Furthermore, Decision Tree-based algorithms were employed for real-time classification [10], optimized for low power and area consumption in a VLSI architecture. The research landscape thus showcases a diverse array of classification techniques, including SVMs, CNNs, MLPs, and Decision Trees, each with its own strengths and trade-offs, contributing to the advancement of early detection and prevention of myocardial infarction through wearable devices.

\section{Hardware used}
Various studies have employed distinct hardware configurations
to achieve optimal performance. Notably, the SmartCardia device[1,2,3], utilizes an ultra-low-power 32-bit microcontroller, 
the STM32L151, featuring an ARM Cortex-M3 operating at a maximum frequency of 32
MHz. This microcontroller is equipped with 48 KB of RAM, 384 KB of Flash memory,
and essential analog peripherals, including a 12-bit ADC. The device also 
incorporates a 710 mAh battery to sustain its operations. Additionally, the 
EFM32 Leopard Gecko development board, detailed in [2], shares a similar 
ARM Cortex-M3 architecture with the SmartCardia INYU, establishing it as a 
relevant low-power target device for experimental purposes. Notably,[1]
and [5] highlight the use of the STM32L151RDT6 microcontroller for real-time 
event-driven classification, coupled with the nRF8001 for Bluetooth Low Energy 
communication. Furthermore, the integration of multiple sensors, such as a 
temperature sensor, heart rate sensor, pressure sensor, and cardiac pode sensor is emphasized in [9].

These sensors gather crucial data for MI detection, which is subsequently 
processed by a computational module.The implementation of the proposed architecture as an Application-Specific Integrated Circuit (ASIC) using SCL180nm Bulk CMOS PDKs and Synopsys design compiler and IC compiler tools is detailed in [10], showcasing a design that occupies 1.38mm2 area and consumes 5.12 $\mu$W power at 1.98V and 8kHz. This comprehensive hardware overview underscores the diverse yet interconnected approaches employed in wearable devices for MI classification, offering valuable insights for researchers and practitioners in the field.

\section{Performance Evaluation}
Several studies have conducted thorough performance evaluations to determine the efficacy of their suggested techniques in the field of wearable myocardial infarction (MI) categorization. For instance, in [1], the authors 
introduced a classification technique exhibiting a remarkable reduction in 
energy consumption by a factor of 3, achieving a medically acceptable accuracy 
level of 90\%. This outperformed previous works, as evidenced by lower 
computational complexity and significantly extended battery 
life. Similarly, [3] presented a methodology 
with an impressive average accuracy of 90.29\%, sensitivity of 90.41\%, and 
specificity of 90.16\% across 10 folds. Additionally, [8] implemented a 
statistical model for early MI detection, yielding a detection rate of 73\% and 
a false alarm rate of 5\% when evaluated on the EDB medical database.

Several papers employed cross-validation schemes, such as 10-fold [2], 5-fold 
[7], and K-fold [9], to rigorously evaluate their models. Notably, [10] 
implemented a classifier using Verilog HDL on an FPGA board, achieving an 
the average sensitivity of 86.18\%, and specificity of 96.5\%. A detailed comparison of various classification methods and their associated accuracies is given in Table 1. Considering the collective findings, it is evident 
that these wearable MI classification technologies, particularly those in [1, 3,
8,10], showcase robust performance metrics encompassing accuracy, sensitivity, 
and specificity, positioning them as promising candidates for further 
exploration and potential integration into wearable healthcare devices.

\section{Conclusion}
The diverse landscape of feature extraction methodologies in wearable Myocardial Infarction (MI) classification, ranging from traditional signal processing to advanced Convolutional Neural Networks (CNNs) and innovative techniques like Hilbert curve mapping, underscores the multidimensional approach to enhancing accuracy and robustness in MI detection. The incorporation of novel features such as normalized signal energy, fractal dimension, and various entropies, as demonstrated in [5], exemplifies the continuous evolution of feature sets to encompass a broader scope of information from ECG signals.

Furthermore, the comprehensive performance evaluations showcased in [1, 3, 8, 10] reveal not only the effectiveness of these methodologies but also their potential for real-world applications. Achieving significant reductions in energy consumption, improved accuracy levels, and extended battery life, these technologies exhibit promising metrics for integration into wearable healthcare devices. The intersection of advanced feature extraction and rigorous performance assessments positions these studies as pivotal contributions to the advancement of MI classification, paving the way for future research and practical implementation in the realm of wearable health technologies.

\end{document}